\documentclass{article}
\usepackage{lipsum}
\usepackage{tikz}
\usepackage{url}
\usepackage{amsfonts}
\usepackage{fancyhdr}
\usepackage{comment}
\usepackage{times}
\usepackage{float}
\usepackage{amsmath}
\usepackage{changepage}
\usepackage{amssymb}
\usepackage{graphicx}
\usepackage{caption}
\usepackage{relsize}

\usepackage{times}
\usepackage{graphicx} 
\usepackage{subfigure} 

\usepackage{natbib}

\usepackage{hyperref}

\usepackage[accepted]{icml2016}

\begin{document} 

\twocolumn[
\icmltitle{Word Embeddings for the Construction Domain}

\icmlauthor{Antoine J.-P. Tixier}{antoine.tixier-1@colorado.edu}
\icmladdress{Postdoctoral Researcher, Computer Science Laboratory, \'{E}cole Polytechnique, Palaiseau, France}
\icmlauthor{Michalis Vazirgiannis}{mvazirg@lix.polytechnique.fr}
\icmladdress{Professor, Computer Science Laboratory, \'{E}cole Polytechnique, Palaiseau, France}
\icmlauthor{Matthew R. Hallowell}{matthew.hallowell@colorado.edu}
\icmladdress{Associate Professor and Beavers Faculty Fellow, Department of Civil Engineering, CU Boulder, USA}

\vskip 0.3in]

\begin{abstract} 
We introduce word vectors for the construction domain. Our vectors were obtained by running {\small\verb|word2vec|} on an 11M-word corpus that we created from scratch by leveraging freely-accessible online sources of construction-related text. We first explore the embedding space and show that our vectors capture meaningful construction-specific concepts. We then evaluate the performance of our vectors against that of ones trained on a 100B-word corpus (Google News) within the framework of an injury report classification task. Without any parameter tuning, our embeddings give competitive results, and outperform the Google News vectors in many cases. Using a keyword-based compression of the reports also leads to a significant speed-up with only a limited loss in performance. We release our corpus and the data set we created for the classification task as publicly available, in the hope that they will be used by future studies for benchmarking and building on our work.
\end{abstract} 
\section{Introduction}
In construction like in many other industries, larger and larger amounts of digital natural text are becoming available. The needs to efficiently process that text are pressing. However, the recent text mining approaches introduced in the construction field do not capitalize on the latest advances in Natural Language Processing (NLP). For instance, the feature extraction system of \cite{tixier2016automated} relies on manually written rules and human-built dictionaries of keywords, while \cite{chokor2016analyzing, williams2014predicting, hsu2013content, caldas2003automating} are all based on the traditional vector space model and Term Frequency - Inverse Document Frequency (TF-IDF) weighting scheme \cite{salton1988term}.

In this paper, we apply word embeddings to the construction domain for the first time. Word embeddings are low-dimensional continuous representations of words that have recently rose to fame following the introduction of the acclaimed {\small\verb|word2vec|} model \cite{mikolov2013efficient}. Provided a large enough corpus, this model was shown to be able to generate embeddings of unmatched quality at minimum cost. These high quality embeddings were in turn shown to boost many NLP tasks like syntactic and semantic word similarity \cite{mikolov2013efficient}, machine translation \cite{mikolov2013exploiting},  or document classification \cite{kusner2015word}.

The contributions of this paper are fourfold :

\begin{itemize}

\item we release an 11M-word corpus of construction-related text that we created from scratch by leveraging only publicly available resources. We believe this corpus to be one of the largest publicly available collections of raw construction-specific text to date,

\item we show that our corpus can be used to learn word embeddings that both encode meaningful construction-specific concepts,

\item we introduce a novel data set for injury report classification (5,845 cases, 11 dependent variables). In addition to document categorization performance, this data set can be used to evaluate keyword extraction performance,

\item using an injury report classification task as a case study, we show that our custom (i.e., local) embeddings outperform in many cases global word vectors learned on a general 100B-word corpus (Google News).

\end{itemize}

The rest of this paper is structured as follows. We first provide an introduction to the concept of word embeddings. Then, we discuss the creation of our construction-specific word vectors and explore the embedding space. Finally, we apply our vectors to the task of injury report classification and quantitatively compare their performance to that of the Google News ones.

\section{Word Embeddings}{\label{sec:embeddings}}

\subsection{Limitations of the vector space model}
Traditionally, text has been represented with the vector space model \cite{salton1988term}. Within this framework, each unique term (i.e., unigram, bigram, etc., up to a certain order) in the universe of documents is considered as an independent dimension of the space, and is encoded as a so-called ``one-hot'' vector. In that discrete space, documents are represented as sparse vectors where the entries are usually binary ($1$ when the word is present in the document, $0$ else), occurrence counts (``bag-of-words'' approach), or TF-IDF weights.

This approach is limited because it considers terms as independent units. Therefore, semantic (meaning) and syntactic (grammar) term-term dependence are completely overlooked. For instance, the word \textit{hammer} may be represented as the vector $[0,0,1,0,...,0,0,0,0]$ and the word \textit{tool} as the vector $[0,0,0,0,...,0,0,1,0]$. It follows that $\overrightarrow{hammer} \cdot \overrightarrow{tool} = 0$. Put differently, according to the vector space model, \textit{hammer} and \textit{tool} are orthogonal. They are as dissimilar as \textit{hammer} and \textit{truck}.

The vector space model also suffers from the \textit{curse of dimensionality}. For instance, if the goal is to model the probability distribution of a sequence of $n=6$ terms in a corpus of vocabulary $V$ of size $|V| = 10^{5}$ unique words, the space that needs to be filled is a $6$-dimensional hypercube where each dimension has $10^{5}$ slots. One would need astronomic amounts of data to perform this task (called $n$-gram language modeling). This is why traditionally, only bigrams and trigrams models have been used in practice \cite{katz1987estimation}. Of course, using such short contexts significantly reduce the amount of term-term relationship information that can be captured.

\subsection{Distributed word representations}
\begin{figure}[h]
\centering
\captionsetup{justification=centering,margin=0.1cm,font=small}
\includegraphics[height=3.5cm, width=8cm]{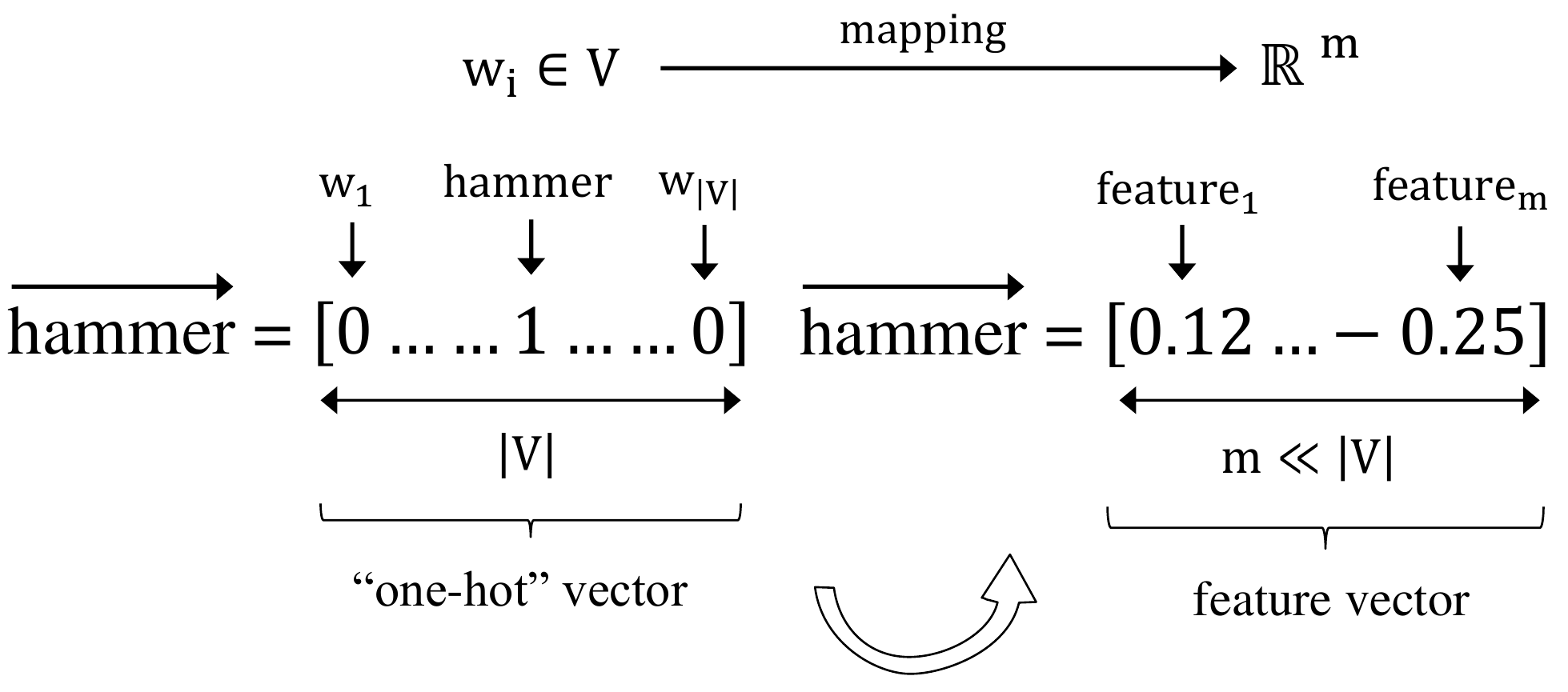}
\caption{$n$-gram model (left) versus word embeddings (right). Where $V$ is the vocabulary of size $\left\vert{V}\right\vert$ (comprising words $w_{1}$ to $w_{\left\vert{V}\right\vert}$), and $m$ is the dimension of the embedding space.}
\label{fig:embed}
\end{figure}

The two aforementioned limitations have motivated the use of \textit{distributed representations of words}, also known as \textit{word embeddings} or \textit{word vectors} \cite{bengio2003neural}. As shown in Figure \ref{fig:embed}, word embeddings map each word in the vocabulary to a real-valued vector in a dense \textit{continuous} space of dimension $m \ll |V|$. The $m$ features encode concepts shared by all words. Typically, while $|V|$ lies in the $[10^{5},10^{6}]$ interval, $m$ takes values within $[100,500]$.

Recently, \cite{mikolov2013efficient} showed that high quality word embeddings could be obtained in a very fast way as a side effect of feeding very large amounts of text to a shallow neural network. The underlying assumption is that a model simple enough to be fed very large amounts of text generates better embeddings than a more complex model that can only afford to be trained on less data. 

Moreover, the embeddings generated after training on large corpora were shown to encode amazingly good syntactic and semantic regularities as simple vector operations \cite{mikolov2013linguistic}. Constant linear translations were shown to capture many concepts like \textit{pluralization}, \textit{gender}, \textit{country-capital}, or \textit{genius-field}. Classical examples include $\overrightarrow{cats} - \overrightarrow{cat} = \overrightarrow{mice} - \overrightarrow{mouse}$, $\overrightarrow{king} - \overrightarrow{man} + \overrightarrow{woman} = \overrightarrow{queen}$, $\overrightarrow{france} - \overrightarrow{paris} = \overrightarrow{italy} - \overrightarrow{rome}$, and $\overrightarrow{einstein} - \overrightarrow{scientist} + \overrightarrow{painter} = \overrightarrow{picasso}$.

\subsection{word2vec}
\begin{figure}[h]
\centering
\captionsetup{justification=centering,margin=0.1cm,font=small}
\includegraphics[height=1.4cm, width=8.5cm]{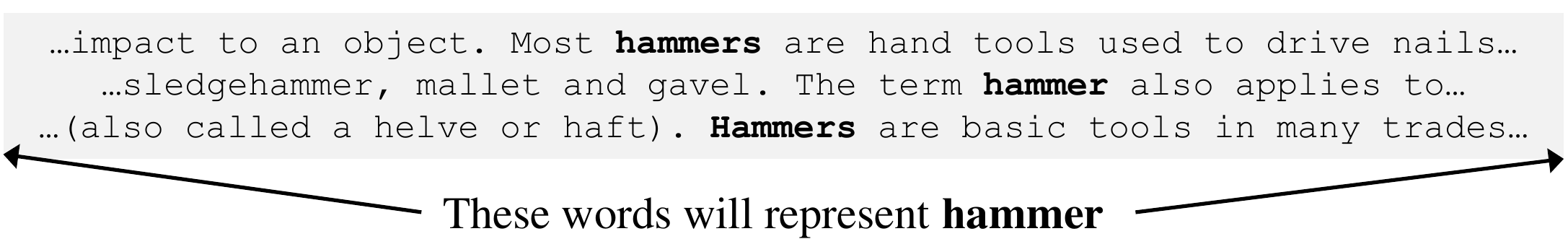}
\caption{Intuition for the Distributional Hypothesis.}
\label{fig:dist}
\end{figure}

The model of \cite{mikolov2013efficient}, also referred to as {\small\verb|word2vec|}, is based on the \textit{Distributional Hypothesis} \cite{harris1954distributional}, which can roughly be summarized as ``we shall know a word by the company it keeps'', and is simply illustrated in Figure \ref{fig:dist}.

\begin{figure}[h]
\centering
\captionsetup{justification=centering,margin=0.1cm,font=small}
\includegraphics[height=6cm, width=5cm]{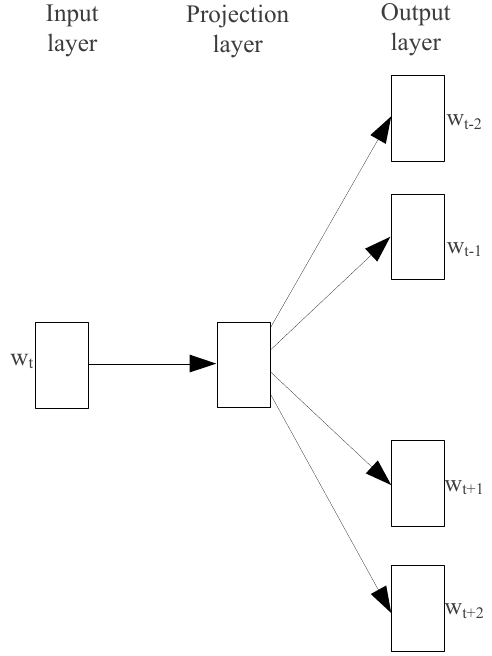}
\caption{Skip-gram architecture with a context of size $4$ ($2$ preceding words, $2$ following words). The task is to predict the surrounding words given the current word $w_{t}$ \cite{mikolov2013exploiting}}
\label{fig:skip}
\end{figure}

More precisely, {\small\verb|word2vec|} first builds $(context, target)$ pairs by linearly parsing the input text from start to finish, where $target$ is a given word and its $context$ of size $2n$ is made of the $n$ preceding and $n$ following words. Reasonable values of $n$ are around $5$, which captures much richer word-word relationship information than $2$-gram or $3$-gram models. {\small\verb|word2vec|} then iteratively passes the $(context, target)$ pairs to a shallow neural network (featuring an input, projection, and output layer only), with the task of predicting either the $target$ word given its $context$ (CBOW architecture) or the $context$ of a given $target$ word (skip-gram architecture), as described in Figure \ref{fig:skip}.

Formally, the objective of the skip-gram model is to maximize:
\begin{equation}
\frac{1}{T}\sum_{t=1}^{T}\sum_{-n \leq j \leq n} \log p(w_{t+j}|w_{t})
\end{equation}
Where $2n$ is the context size, and the training corpus is a sequence of words $w_{1}, w_{2}, ..., w_{T}$. Note that to obtain good results we must have $T \gg |V|$.

At each iteration, the error gets backpropagated via stochastic gradient descent and accordingly updates the weights of the projection-output and input-projection matrices. Those weight matrices precisely contain the desired word vectors. They are initialized randomly (i.e., words are initially dispersed at random within the space), and then, as training takes place, words get closer to each other (or more distant from each other) as a reaction to the error, in an attraction-repulsion spring-like fashion \cite{rong2014word2vec}. The movements are coarse at first and get finer and finer as the neural network gets better and better and the error diminishes. Upon several epochs of training with a decreasing learning rate, the final vectors are ready. 

Within the resulting embedding space, words sharing similar meanings occupy neighboring positions. While both the input and the output weight matrices can theoretically be used or combined, the $|V|$ by $m$ input matrix is generally used as the word vectors \cite{mikolov2013linguistic}.

Several facts are interesting to note:
\begin{itemize}

\item the word vectors are obtained as a \textit{side effect} of training. The neural network is actually never used for the prediction task it was trained to perform,

\item thanks to its simplicity (no hidden layer, only a linear projection layer), {\small\verb|word2vec|} is extremely fast to train. Using the original C code with multiprocessing has been reported to offer training rates of several billions of words per hour \cite{mikolov2013exploiting},

\item because the process is stochastic, running {\small\verb|word2vec|} twice on the same corpus will not return the same word vectors. However, the similarities (i.e., distances) between words and the organization of concepts remain the same,

\item {\small\verb|word2vec|} was released as an open source project\footnote{\url{https://code.google.com/archive/p/word2vec/}} by Google in 2013 and has since then been ported to other languages. In this study, we used the popular {\small\verb|gensim|} Python implementation\footnote{\url{https://radimrehurek.com/gensim/models/word2vec.html}} \cite{gensim}.

\end{itemize}

\section{Word Vectors Creation}{\label{sec:corpus}}
The main objective at this stage was to gather as much construction-related text as possible. Indeed, {\small\verb|word2vec|} requires large quantities of text to yield good embeddings (the more the better, with diminishing returns). However, although the size of our input text was important, we needed that text to be related to construction as much as possible. In what follows, we present, sorted by decreasing size, the various publicly available resources we leveraged to create our construction-specific corpus.

\subsection{Wikipedia}
A first and obvious large source of text was the English Wikipedia. We selected all pages related to the \textit{Construction} category, and all their children and grandchildren, resulting in a final list of 12,256 pages\footnote{\url{https://github.com/Tixierae/WECD/blob/master/list_wikipedia_pages.txt}}. We then used Wikipedia's {\small\verb|Special:Export|} tool\footnote{\url{https://en.wikipedia.org/wiki/Special:Export}} to download all these pages as XML files. All the text corresponding to the content of these files was then extracted, yielding a corpus of 6,383,953 words (403,763 unique words).

\subsection{ELCOSH}
The Electronic Library of Construction Occupational Safety and Health\footnote{\url{http://www.elcosh.org/index.php}} (ELCOSH) also turned out to be a valuable source of construction-related text. We scraped the pages related to \textit{Handouts} (245 documents), \textit{Toolbox Talks} (179), \textit{Research Reports} (166), and \textit{Training Materials} (102). Whenever text was not directly available on the web page, we parsed the linked PDF document(s). We thus obtained a corpus of 2,074,769 words (56,070 unique words).

\subsection{OSHA} \label{sub:osha}
We also extracted text from the Occupational Safety and Health Organization (OSHA) website, as follows: \\
\textbf{IMIS}. We queried the Integrated Management Information System (IMIS) accident search tool\footnote{\url{https://www.osha.gov/pls/imis/accidentsearch.html}} for reports pertaining to the NAICS codes 236, 237, and 238. These codes respectively correspond to the \textit{Construction of Buildings}, \textit{Heavy and Civil Engineering}, and \textit{Specialty Trade Contractors} categories, and were associated at the time of the study with 2,691, 2,430, and 9,780 injury reports respectively. Altogether, the 14,901 reports returned a corpus of 1,497,056 words (25,382 unique words). \\
\textbf{SLTC}. We also leveraged the list of Safety and Health Topics\footnote{\url{https://www.osha.gov/SLTC/text_index.html}}. 165 pages were parsed in total, giving a 37,629-word corpus (4,851 unique words).

\subsection{CPWR}
\textbf{Publications}. The Center for Construction Research and Training, also known as CPWR, offers a list of research findings and articles on the \textit{publications} page of its website\footnote{\url{http://www.cpwr.com/publications/publications}}. More precisely, we parsed the research reports found in the \textit{Design for Safety}, \textit{Accident Data Analysis}, \textit{Health Hazards}, \textit{Safety Hazards}, and \textit{Hispanic Workers} pages. We also parsed the PDF documents linked in the \textit{Key Findings from Research} webpage. This made for a list of 162 documents in all, yielding a corpus of 416,150 words (26,095 unique words). \\
\textbf{Workbook}. We also parsed the \textit{Day Laborers's Health and Safety Workbook}\footnote{\url{http://www.cpwr.com/sites/default/files/publications/DayLaborersTrainingGuide-UIC-edition-English.pdf}}. This 453-page document returned a 68,776-word (6,679 unique words) corpus.

\subsection{NIOSH FACE}
Another source of text specific to the construction domain was found in the form of accident reports from the National Institute for Occupational Safety and Health (NIOSH) Fatality Assessment and Control Evaluation (FACE) program\footnote{\url{https://www.cdc.gov/niosh/face/inhouse.html}}. The text from the 249 reports belonging to the \textit{Construction} category (at the time of the study) was gathered. Whenever the webpages included links to PDF documents, those documents were also parsed. This eventually gave us a corpus of 381,969 words (13,770 unique words).

\subsection{USACE}
Finally, we parsed the US Army Corps of Engineers (USACE) manual which prescribes the Safety and Health requirements for all Corps of Engineers operations. This 977-page document returned a corpus of 185,449 words (11,621 unique words).

\subsection{Aggregation and learning}
After final cleaning, we obtained an overall corpus of 11,043,511 words (456,402 unique words, 70MB in size), which was split into 55,495 200-word chunks before being passed to {\small\verb|gensim|}. To enable future studies to build on our work, we make our corpus and the different sub-corpora previously presented freely available for download\footnote{\url{https://github.com/Tixierae/WECD/blob/master/corpora.zip}}. 

To generate our word embeddings, we used the Skip-gram architecture, as it was reported to give better performance on small corpora \cite{mikolov2013exploiting}, with standard parameter values: context of size 10, $m=300$, a downsampling threhsold of $10^{-5}$ for high-frequency words, a negative sampling value of 3, and 10 training epochs. Moreover, words that occurred less than five times in the corpus, standard English stopwords, and custom stopwords\footnote{\url{https://github.com/Tixierae/WECD/blob/master/custom_stpwds.txt}} were not embedded. Using multiprocessing, training took only a few minutes on a standard laptop with four virtual cores. The final embeddings had dimensions 32,689 by 300, which respectively correspond to the vocabulary size and to the number of features ($m$).

\section{Exploration of the Embedding Space}{\label{sec:exploration}}
We qualitatively evaluated the quality of our word vectors by assessing the extent to which they encoded meaningful relationships between construction-specific concepts. To do so, we assigned our model different problems, which are next presented (along with the results) by increasing order of complexity.

\begin{figure}[h]
\centering
\captionsetup{justification=centering,margin=0.1cm,font=small}
\includegraphics[height=6cm, width=7cm]{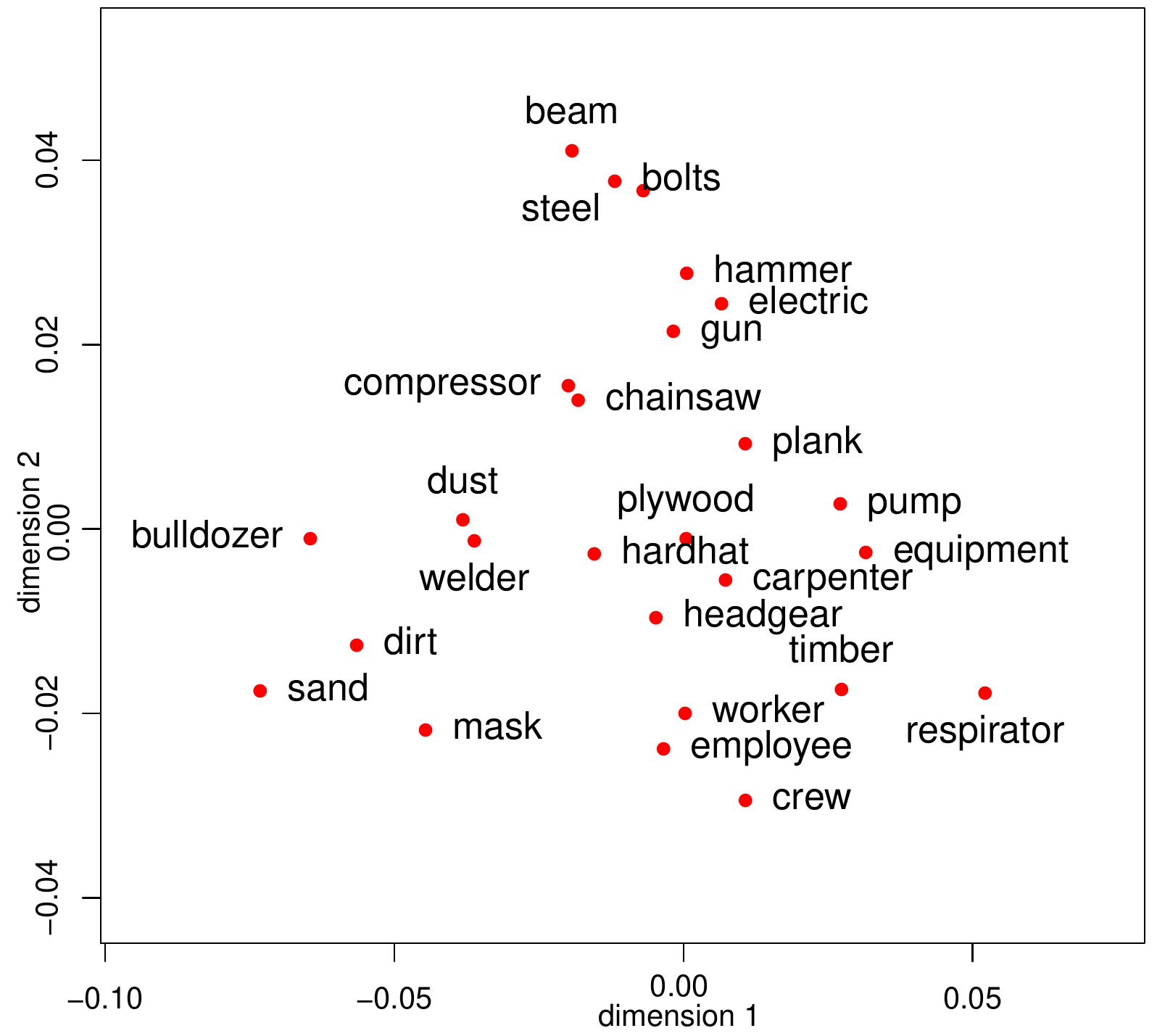}
\caption{Semantically close words occupy neighboring positions in the embedding space.}
\label{fig:clust}
\end{figure}

\textbf{Clustering task}.
We first verified that words similar in meaning were close from one another in our embedding space. To be able to visualize the groupings, we reduced dimensionality with PCA using R \cite{r}. Figure \ref{fig:clust} shows the projections of some selected words onto the first two principal directions. Even if the picture is partial and compressed, we can see that semantically close words, such as ``worker'', ``employee'', and ``crew'', or ``sand'' and ``dirt', indeed occupy neighboring positions. More generally, words that are often found in similar contexts are also close together (e.g., ``bolts'', ``beam'', and ``steel'', ``dust'' and ``welder'').

\textbf{Word similarity task}.
Another way to assess the extent to which our embeddings learned meaningful construction semantics was to ask our model to return the $10$ words closest to a given query (in terms of cosine similarity). 

The cosine of the angle between two vectors $\vec{v_1}$ and $\vec{v_2}$ is a widely-used similarity metric in vector spaces. It is defined as follows: 
\begin{equation}
 cos(\vec{v_1},\vec{v_2})=\frac{\vec{v_1} \cdot \vec{v_2}}{\|\vec{v_1}\| \times \|\vec{v_2}\|}
\end{equation}
where $\|\cdot\|$ is the euclidean distance. A null cosine indicates orthogonality, while values approaching the unit signify that the two vectors are highly similar.

As shown in Tables \ref{table:acid} to \ref{table:autodesk}, the responses to the queries make a lot of sense. Despite the fact that our model was fed raw text in a fully unsupervised fashion (that is, it was not passed any annotated lists of construction-specific words), it managed to learn about different kinds of acids, construction materials and personnel, as well as Computer Aided Design (CAD)/Building Information Modeling (BIM) tools (among other things).
\begin{table}[!h]
\small
\centering
\begin{tabular}{lr}
 \hline
 words closest to ``\textit{acid}'' & \textit{d}\\ 
\hline
\hline
 sulfuric     & .89 \\
 hydrochloric & .85 \\
 ammonium     & .83 \\
 hydroxide    & .82 \\
 hydrofluoric & .81 \\
 dissolves    & .81 \\
 potassium    & .81 \\
 phosphoric   & .81 \\
 chlorine     & .81 \\
 oxidizes     & .81 \\
\hline
\end{tabular}
\captionsetup{justification=centering, size=scriptsize}
\caption{top 10 words closest to ``acid'' (cosine distance).}
\label{table:acid}
\end{table}
\begin{table}[!h]
\small
\centering
\begin{tabular}{lr}
 \hline
 words closest to ``\textit{2x4}'' & \textit{d}\\ 
\hline
\hline
 2-foot      & .903 \\
 two-by-four & .901 \\
 34-inch     & .901 \\
 16-ft-long  & .898  \\
 dunnage     & .896 \\
 4x8         & .896  \\
 2-by-4      & .892 \\
 2-by-4s     & .892 \\
 5-in        & .892 \\
 lathing     & .892 \\
\hline
\end{tabular}
\captionsetup{justification=centering, size=scriptsize}
\caption{top 10 words closest to ``2x4'' (cosine distance). ``2x4'' is a standard dimension of rough planks.}
\label{table:2x4}
\end{table}
\begin{table}[!h]
\small
\centering
\begin{tabular}{lr}
 \hline
 words closest to ``\textit{foreman}'' & \textit{d}\\ 
\hline
\hline
 supervisor     & .85 \\
 coworker       & .82 \\
 superintendent & .82 \\
 coworkers      & .82 \\
 leadman        & .80 \\
 groundman      & .79 \\
 instructed     & .79 \\
 ppat           & .79  \\
 employee       & .78 \\
 crew           & .78  \\
\hline
\end{tabular}
\captionsetup{justification=centering, size=scriptsize}
\caption{top 10 words closest to ``foreman'' (cosine distance).}
\label{table:foreman}
\end{table}
\begin{table}[!h]
\small
\centering
\begin{tabular}{lr}
 \hline
 words closest to ``\textit{autodesk}'' & \textit{d}\\ 
\hline
\hline
 autocad      & .91 \\
 nx           & .88 \\
 microstation & .88 \\
 navisworks   & .88 \\
 zwcad        & .88 \\
 revit        & .88 \\
 graphisoft   & .88 \\
 c3d          & .87 \\
 ironcad      & .86 \\
 plm          & .86 \\
\hline
\end{tabular}
\captionsetup{justification=centering, size=scriptsize}
\caption{top 10 words closest to ``autodesk'' (cosine distance).}
\label{table:autodesk}
\end{table}

Without having been taught technical construction jargon or word pronunciation, the model knows for instance that the abbreviation ``2x4'', which refers to standard dimensions of rough planks, can also be written ``two-by-four''. It also knows that a foreman is a supervisor, and that Microstation and ZWCAD are competitors of Autocad (Autodesk's flagship software) in the CAD/BIM market. This approach could be used to automatically build dictionaries of synonyms that would improve the flexibility of tools such as that of \cite{tixier2016automated}.

\textbf{Word mismatch task}. We also tested whether our word vectors could discriminate between different construction-specific concepts. To do so, we used the \textit{``which words does not match?''} task. This problem consists in identifying the word that should be removed from a given list because it is semantically unrelated to the other words. To solve it, the system computes the cosine similarity in the embedding space between each word in the list and all the others, and averages the results. The word that does not match is the one that is associated with the lowest score.

As shown in Table \ref{table:mismatch}, our model can successfully differentiate between the construction \textit{material}, \textit{trades}, \textit{software}, \textit{equipment}, \textit{tools}, and \textit{injury} concepts (among others).

\begin{table}[h]
\small
\centering
\begin{tabular}{c|c}
\hline
which word does not match? & response \\
\hline
\hline
pipe \textbf{roof} trench cables ground & \textit{roof}\\
asbestos silica \textbf{rebar} fiberglass dust & \textit{rebar}\\
carpenter employee laborer electrician \textbf{bim} & \textit{bim}\\
car truck \textbf{drill} excavator manlift crane & \textit{drill}\\
hernia \textbf{building} injury fracture burn sprain & \textit{building}\\
\hline
\end{tabular}
\captionsetup{justification=centering, size=scriptsize}
\caption{Thanks to word embeddings, a machine can differentiate between construction-specific concepts, despite never having been instructed about them in a supervised way.}
\label{table:mismatch}
\end{table}

\textbf{Word analogy task}. While the previous problem was a simple mismatch detection task, finding word analogies is slightly more complex. This problem can be stated as finding the answer to ``a'' is to ``b'' as ``c'' is to ``unknown''. The solution can be found via the \textit{vector offset method} \cite{mikolov2013linguistic}. We simply compute $\vec{b}-\vec{a}+\vec{c}$ and retrieve the word whose vector is the closest (in terms of cosine similarity) from the output. The results for this job are shown in Table \ref{table:analogy}, and are very convincing. Again, the implication is that following a phase of \textit{fully unsupervised} training on \textit{raw unannotated} construction-related text, {\small\verb|word2vec|} was able to learn construction-specific concepts and organize them meaningfully within the embedding space.

\begin{table}[h]
\small
\centering
\begin{tabular}{c|c|c|c}
\hline
...  & is to ... & as ...& is to: \textit{response}\\
\hline
\hline
brick & mason & wood & \textit{carpenter} \\
volvo & trucks & autodesk & \textit{autocad} \\
beams & ironworker & concrete & \textit{finisher} \\
nails & hammer & bolts & \textit{wrench} \\
\hline
\end{tabular}
\captionsetup{justification=centering, size=scriptsize}
\caption{Our word embeddings can perform simple analogy tasks successfully.}
\label{table:analogy}
\end{table}

\begin{figure}[ht]
\centering
\captionsetup{justification=centering,margin=0.1cm,font=small}
\includegraphics[height=6cm, width=7.5cm]{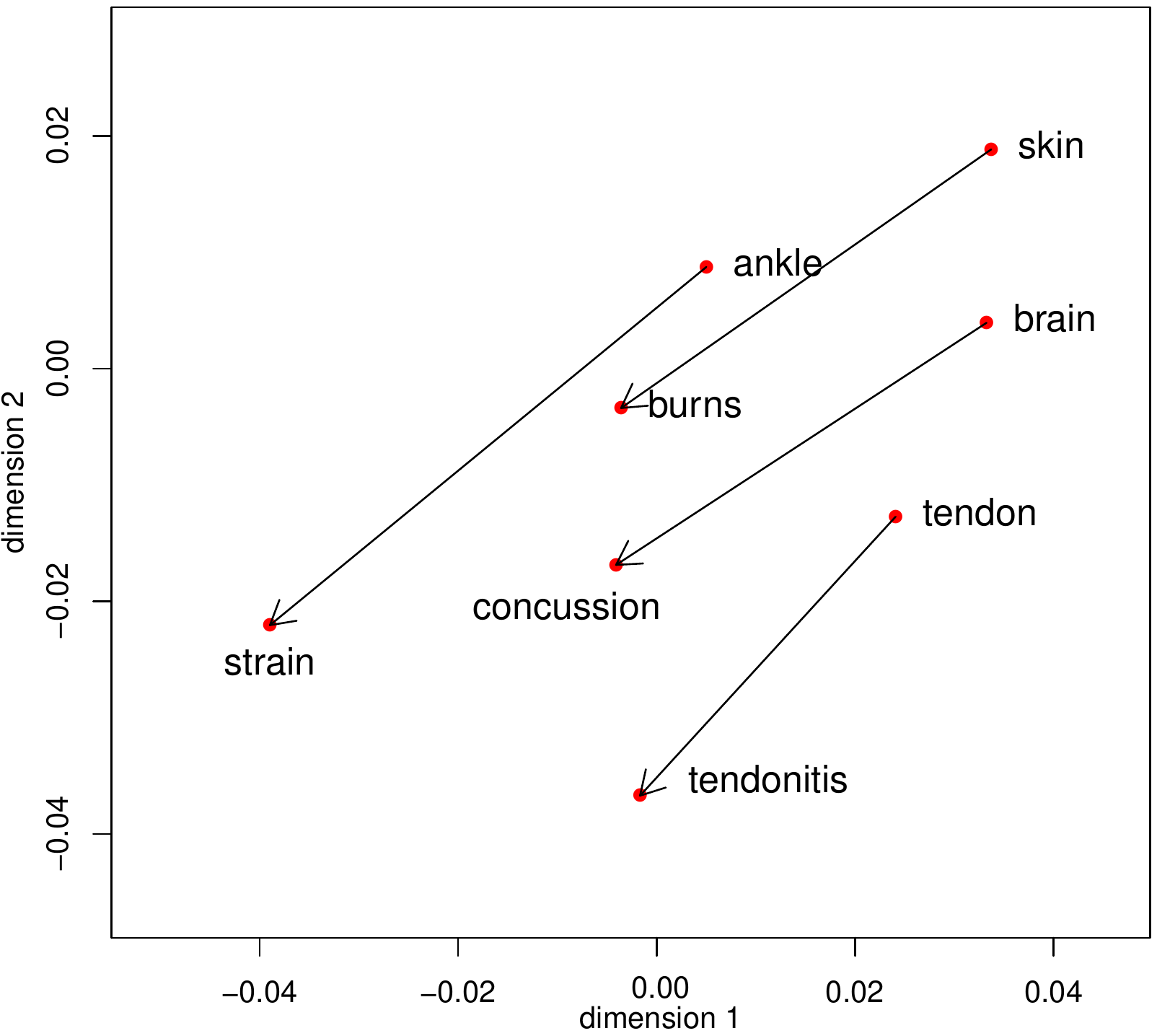}
\caption{A simple constant linear translation links \textit{body parts} to sustained \textit{injuries}.}
\label{fig:low_plot_1}
\end{figure}

\begin{figure}[ht]
\centering
\captionsetup{justification=centering,margin=0.1cm,font=small}
\includegraphics[height=6cm, width=7.5cm]{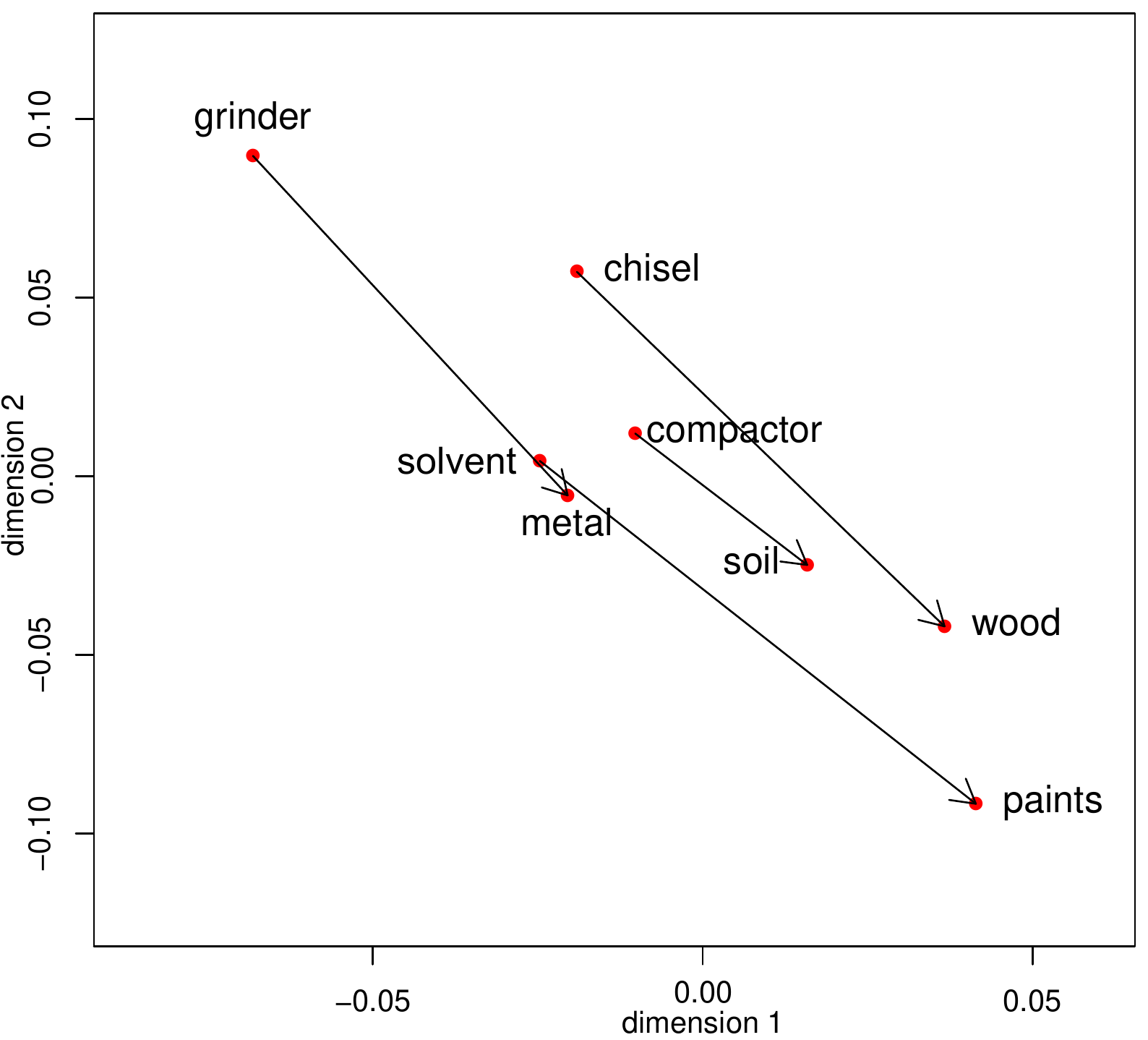}
\caption{A roughly constant linear translation allows one to go from a \textit{tool} or \textit{equipment} to the corresponding \textit{material}.}
\label{fig:low_plot_2}
\end{figure}

\begin{figure}[ht]
\centering
\captionsetup{justification=centering,margin=0.1cm,font=small}
\includegraphics[height=6cm, width=7.5cm]{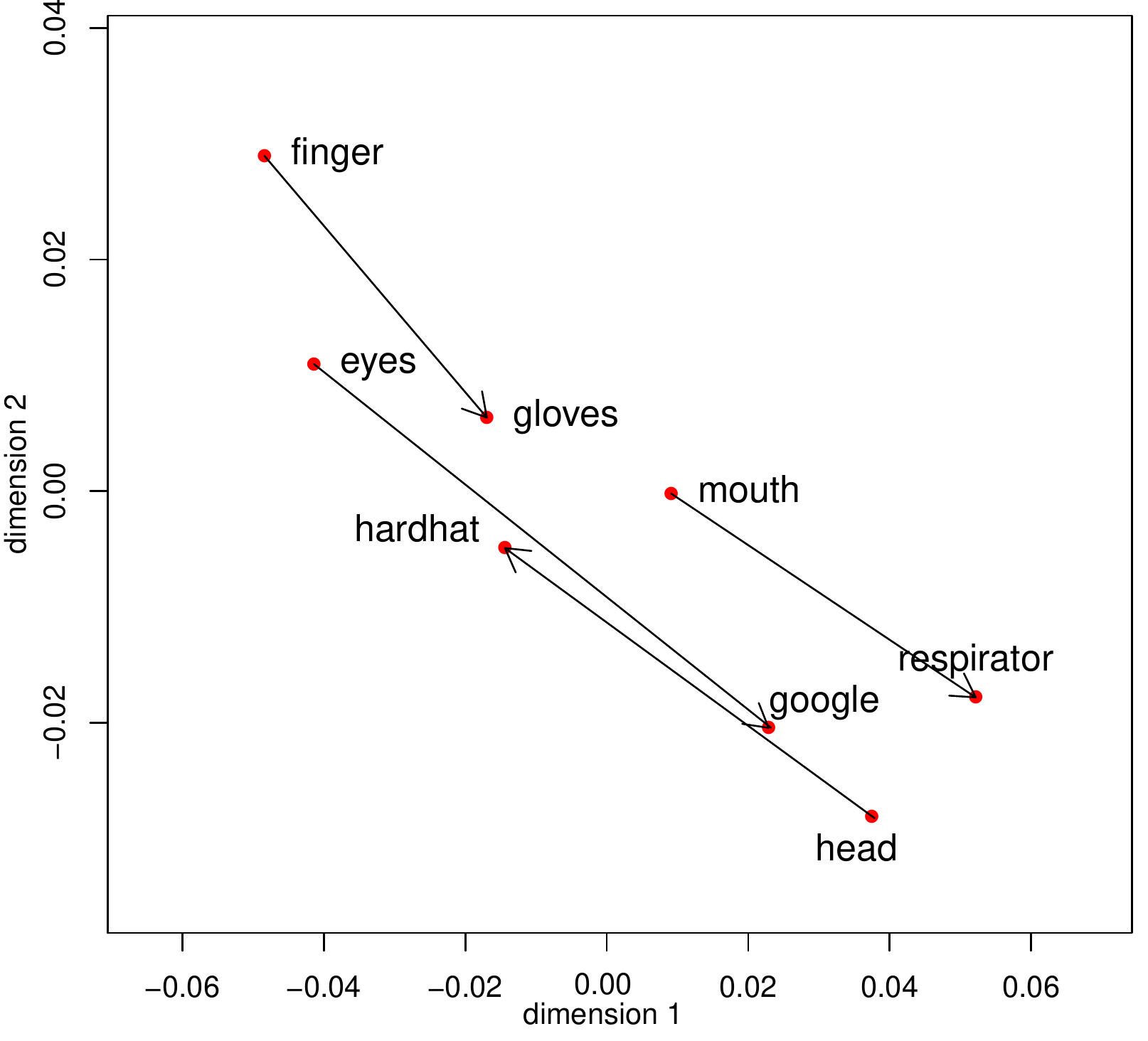}
\caption{\textit{Body parts} and their respective \textit{PPE} are mapped by an approximately constant translation. Interestingly, head-hardhat and eyes-google have same magnitude, but opposite directions.}
\label{fig:low_plot_3}
\end{figure}

\begin{figure}[ht]
\centering
\captionsetup{justification=centering,margin=0.1cm,font=small}
\includegraphics[height=6cm, width=7.5cm]{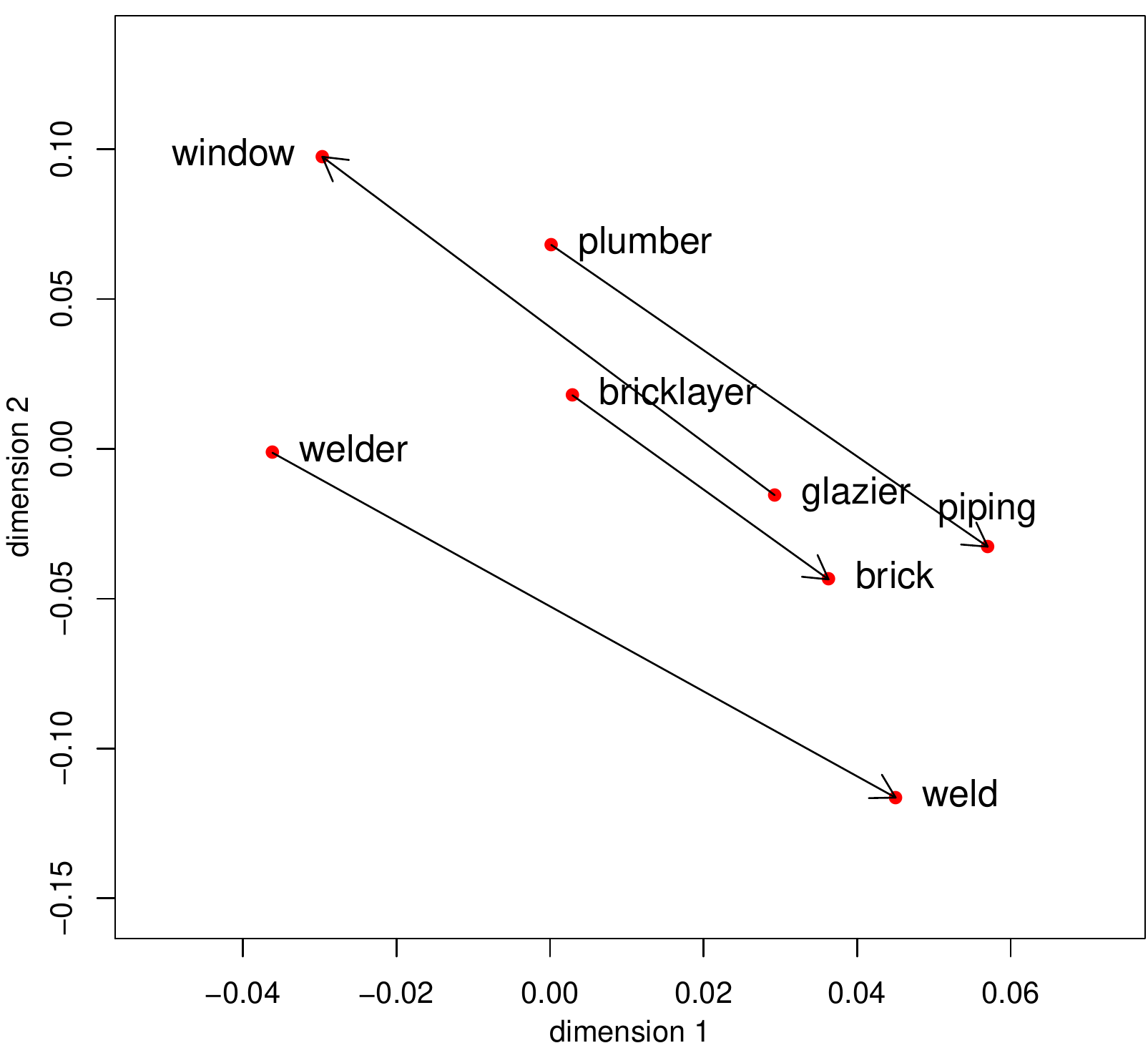}
\caption{The concept \textit{trade}-\textit{material} is encoded by a roughly constant linear translation. welder-weld and glazier-window have same magnitude but opposite directions.}
\label{fig:low_plot_4}
\end{figure}

\textbf{Visualizing analogies}. As already explained, concepts are implicitly mapped in a way that allows their relationship to be accessed via simple vector operations, such as linear translations. We decided to visualize some of these translations in the PCA space (see Figures \ref{fig:low_plot_1} to \ref{fig:low_plot_4}). Even though plotting the vectors on the first two principal directions gives a compressed and incomplete representation of what actually happens in the full embedding space, we still can observe that our word vectors encode meaningful regularities. For instance, roughly constant mappings exist between \textit{body parts} and the corresponding \textit{injuries} or \textit{Personal Protective Equipment} (\textit{PPE}), and between \textit{materials} and the associated \textit{tools} or \textit{trade}.

\section{Application to Injury Report Classification}{\label{sec:application}}
We also wanted to quantitatively evaluate the quality of our word vectors in a real-life application. To proceed, we designed an \textit{injury report classification} task. In what follows, we describe the creation of our data set, the experimental set-up, and the results.

\subsection{Data set creation}\label{sub:data}
To create our data set, we iterated through the OSHA IMIS injury reports (for the NAICS codes 236, 237, and 238, see subsection \ref{sub:osha}) and retained the reports that were associated with a ``complete'' table, that is, a table featuring the following seventeen fields (regardless of whether the fields were blank or not): (1) \textit{identification number}, (2) \textit{keywords}, (3) \textit{narrative}, (4) \textit{end use}, (5) \textit{project type}, (6) \textit{project cost}, (7) \textit{number of stories}, (8) \textit{building height}, (9) \textit{fatality} (yes/no answer), (10) \textit{number of employees involved}, (11) \textit{inspection number}, (12) \textit{age}, (13) \textit{gender}, (14) \textit{injury severity} (called \textit{degree} on the OSHA website), (15) \textit{injury type} (\textit{nature}), (16) \textit{trade} (\textit{occupation}), and (17) \textit{additional information}. Other than making easier the systematic and automated creation of the data set, we proceeded in such a manner because we assumed that the reports associated with well-formed tables would be of better quality.

We were thus able to obtain a structured data set of 5,845 rows (reports) by 18 columns (NAICS code plus all the seventeen aforementioned field names). Because the \textit{age} and \textit{gender} columns had blank fields only, we removed them, making for a final count of 16 columns, including 11 potential dependent variables. For reproducibility, and also so that future studies can use it for benchmarking, we release our data set as publicly available\footnote{\url{https://github.com/Tixierae/WECD/blob/master/classification_data_set.csv}}. It can be used to evaluate injury report classification performance (for various categories), but also keyword extraction performance. 

All of the 5,845 reports had their major fields (\textit{narrative}, \textit{id number}...) filled out. However, we decided to classify reports for \textit{injury severity}, \textit{injury type}, and \textit{trade}. We thus had to retain only the reports for which the fields corresponding to these categories had not been left blank. This yielded a final subset of 1,688 reports ($0$-based index provided here\footnote{\url{https://github.com/Tixierae/WECD/blob/master/index_overlap.txt}}) that was still big enough to suit our experiments.

\subsection{$k$-nearest neighbor classifier}\label{sub:knn}
$k$-nearest neighbor \cite{cover1967nearest} is a basic machine learning algorithm that is not \textit{supervised} nor \textit{unsupervised}. Indeed, while it does require a set of ``training'' data, the $k$-nearest neighbor classifier does not derive rules from them like Support Vector Machines or Random Forest. It simply computes the \textit{distance} (in the feature space) between a new observation and each observation in the training set, and aggregates the target values of the $k$ neighbors closest to the new observation to generate a prediction for the target value of the new observation. In the case of a \textit{continuous} dependent variable, the \textit{mean} or \textit{median} of the neighbors' target values is returned, whereas in the \textit{categorical} case, the most frequent class is used.

Because it waits until a prediction is required to actually process the training data, the $k$-nearest neighbor algorithm is said to be a \textit{lazy} learning technique. Determining $k$ is subject to the classical \textit{bias-variance} trade-off: considering only a few neighbors tends to overfit the data (high variance-low bias), whereas taking too many neighbors into account underfits them (low variance, but high bias). Tuning this parameter is therefore important.

\subsection{Word Mover's Distance}
As was previously explained, the $k$-nearest neighbor classifier requires a \textit{distance} to compute similarity between observations in the feature space. For this purpose we used the recently introduced Word Mover's Distance \cite{kusner2015word}, or WMD. It provides an intuitive way to convert word embeddings to document distances.

The WMD is a simple adaptation of the Earth Mover's Distance \cite{rubner2000earth}, a well-known metric in the Computer Vision field, to NLP. It measures the distance between two pieces of text $\vec{p_{1}}$ and $\vec{p_{1}}$ as the minimum weighted cumulative cost needed for all words of $\vec{p_{1}}$ to ``travel'' to $\vec{p_{2}}$. It is formally defined as a transportation problem:
\begin{equation}
\textnormal{WMD}(\vec{p_{1}},\vec{p_{2}})=\min\sum_{i,j=1}^{|V|} T_{i,j}c_{i,j}
\end{equation}
where $|V|$ is the size of the vocabulary, $T_{i,j}$ is the $(i,j)^{th}$ entry of a non-negative transfer matrix $T \in \mathbb{R}^{|V|^2}$ that represents the amount of the $i^{th}$ word in $\vec{p_{1}}$ ($\vec{w_{i}}$) that travels to the $j^{th}$ word in $\vec{p_{2}}$ ($\vec{w_{j}}$), and $c_{i,j}$ is the cost of traveling from $\vec{w_{i}}$ to $\vec{w_{j}}$. This problem is subject to the constraint that $\vec{p_{1}}$ must be entirely transported to $\vec{p_{2}}$.

Furthermore, \cite{kusner2015word} represent a document $\vec{p_{k}} \in \mathbb{R}^{|V|}$ as a normalized vector of word counts in the vector space model (normalized bag-of-words representation), with elements:

\begin{equation}
p_{k_{i}} = \frac{\mathrm{count}(w_{i} \in p_{k})}{\sum_{j=1}^{|V|} \mathrm{count}(w_{j} \in p_{k})}, \forall i \in \{1,...,|V|\}
\end{equation}

What actually makes the connection with word embeddings, is that the travel cost $c_{i,j}$ between $\vec{w_{i}}$ and $\vec{w_{j}}$ is defined as the euclidean distance between them in the word embedding space:

\begin{equation}
c_{i,j}=\|\vec{w_{i}}-\vec{w_{j}}\|
\end{equation}

Thanks to this meaningful travel cost, the WMD provides a natural transition from word embeddings to document distances. An illustrative example is provided in Table \ref{table:wmd}, where the top 5 injury reports closest to a given report are shown. We can see that the neighbors retrieved by the WMD are very relevant to the query in several aspects. First, all neighbors share the same outcome as the query (\textit{hospitalized injury}), second, they all deal (except the closest one), with \textit{fall from height} injuries (like the query). Using in that case the $5$-nearest neighbors would therefore yield accurate forecasts for \textit{injury severity} and \textit{injury type}. 

Apart from plain overlap, some words (shown in \textit{italic} in Table \ref{table:wmd}) like ``initiated'' and ``initial'', ``plant'' and ``facility'', ``knee'' and ``leg'', etc. are very close to each other in the embedding space (due to their synonymy) which even more reduces the WMD between the responses and the queries. It is also interesting to note that the responses and the query are very similar in size. Having documents of roughly equal size thus must ease the transportation problem and reduce the WMD distance between them.

\subsection{Experimental set-up}
\textbf{Benchmarking}. We compared the quality of our custom construction-specific word vectors against that of general embeddings\footnote{\url{https://drive.google.com/file/d/0B7XkCwpI5KDYNlNUTTlSS21pQmM/edit}} trained on a 100B-word Google News corpus \cite{mikolov2013distributed}. These word vectors were generated using the same parameters as we used in our study ($m=300$, subsampling threshold of $10^{-5}$, negative sampling of 3...), but with the continuous bag-of-words (CBOW) architecture. Recall that in our case, we used the Skip-gram architecture as it is held to perform better on small corpora. While Google News vocabulary size is 3 million, we only selected the vectors of the 32,689 words we had custom embeddings for.

\textbf{$4$-fold cross validation}.
We used the data set of 1,688 reports described in subsection \ref{sub:data} with three prediction tasks: \textit{injury severity}, \textit{injury type}, and \textit{trade}. The number of categories into which observations could fold were $3$, $9$, and $4$, respectively for each task. The words for which no embedding was available were removed from the reports beforehand, as well as punctuation marks. The processed reports can be found here\footnote{\url{https://github.com/Tixierae/WECD/blob/master/reports_processed.txt}} (one report per line). \\
Evaluation was conducted under a $4$-fold cross validation setting: the full set of reports was split into four folds of equal size (422 reports). Then, at each of four steps, three folds were used as the training instances and the last fold was used as the testing set. The $k$-nearest neighbor classifier described in subsection \ref{sub:knn} then generated predictions for each element in the testing set with $k$ varying from $5$ to $25$, by steps of $5$, using both our custom embeddings (custom in what follows) and the Google News embeddings (google in what follows). 

\begin{table*}[ht]
\small
\centering
\begin{tabular}{|p{14cm}|p{0.7cm}|}
 \hline
reports closest to: ``\textbf{industrial engaged} forming \textbf{treatment} \textit{plant} \textbf{fell} pit \textbf{transported \textit{hospital admitted}} \textbf{treated} dislocated \textit{knee initiated} report ongoing'' & WMD\\ 
\hline
\hline
  foreman \textbf{engaged} bituminous concrete placement chop cut lower services \textbf{transported admitted treated} included surgery \textit{initial} \textbf{treatment} released \textbf{hospital} manager
    & 2.15 \\ 
    \hline
 full regular \textbf{industrial fell} mezzanine \textbf{transported admitted} diagnosed multiple body system fractures transferred \textit{facility} discharged & 2.21 \\
      \hline
    base digging footers seawall part cliff \textbf{fell} engulfing trapped significant period finally extricated \textbf{transported} local \textbf{hospital} \textit{admission} \textbf{treatment} injuries \textit{hospitalized} \textbf{treated} bruised \textit{leg} & 2.22 \\
    \hline
 carpenter \textbf{engaged} interior carpentry commercial \textbf{fell} ceiling joist \textit{fall} height feet services \textbf{transported hospital admitted treated} back neck & 2.24 \\
\hline
layer roof top \textbf{engaged} framing level roof walking stepped cover roof opening \textbf{fell} sustained fractured femur fractures wrists \textbf{transported hospital admitted treatment} & 2.25
\\
\hline

\end{tabular}
\captionsetup{justification=centering, size=scriptsize}
\caption{Top 5 reports closest to a given report. Due to space issues, the processed versions of the reports are shown, where only words that have custom embeddings available have been retained. The words that are both present in the response and the query have been highlighted in \textbf{bold}, while synonyms are shown in \textit{italic}.}
\label{table:wmd}
\end{table*}

\textbf{Bag-of-words baseline}. For comparison purposes, we also used as a baseline the euclidean distance between bag-of-words representations of injury reports (bow in what follows). It was implemented using the {\small\verb|euclidean_distances|} and {\small\verb|CountVectorizer|} functions of the {\small\verb|scikit-learn|} Python library \cite{scikit-learn}.

\textbf{Keyword-based compression}.
The $k$-nearest neighbor algorithm has to compute the distance between every element of the test set and every element of the training set to sort the observations and select the nearest ones to generate predictions. This represents a significant number of distances to compute within each fold, which takes time because the WMD is an expensive metric.

To speed-up the process, we experimented with a keyword-based compression of injury reports. That is, rather than using the full text of the reports, we tried to represent them with their keywords. The distances (WMD or euclidean) were then computed for these compressed representations rather than based on the full text.

\begin{figure*}[ht!]
\centering
\captionsetup{justification=centering,margin=0.1cm,font=small}
\includegraphics[height=8.5cm, width=14.5cm]{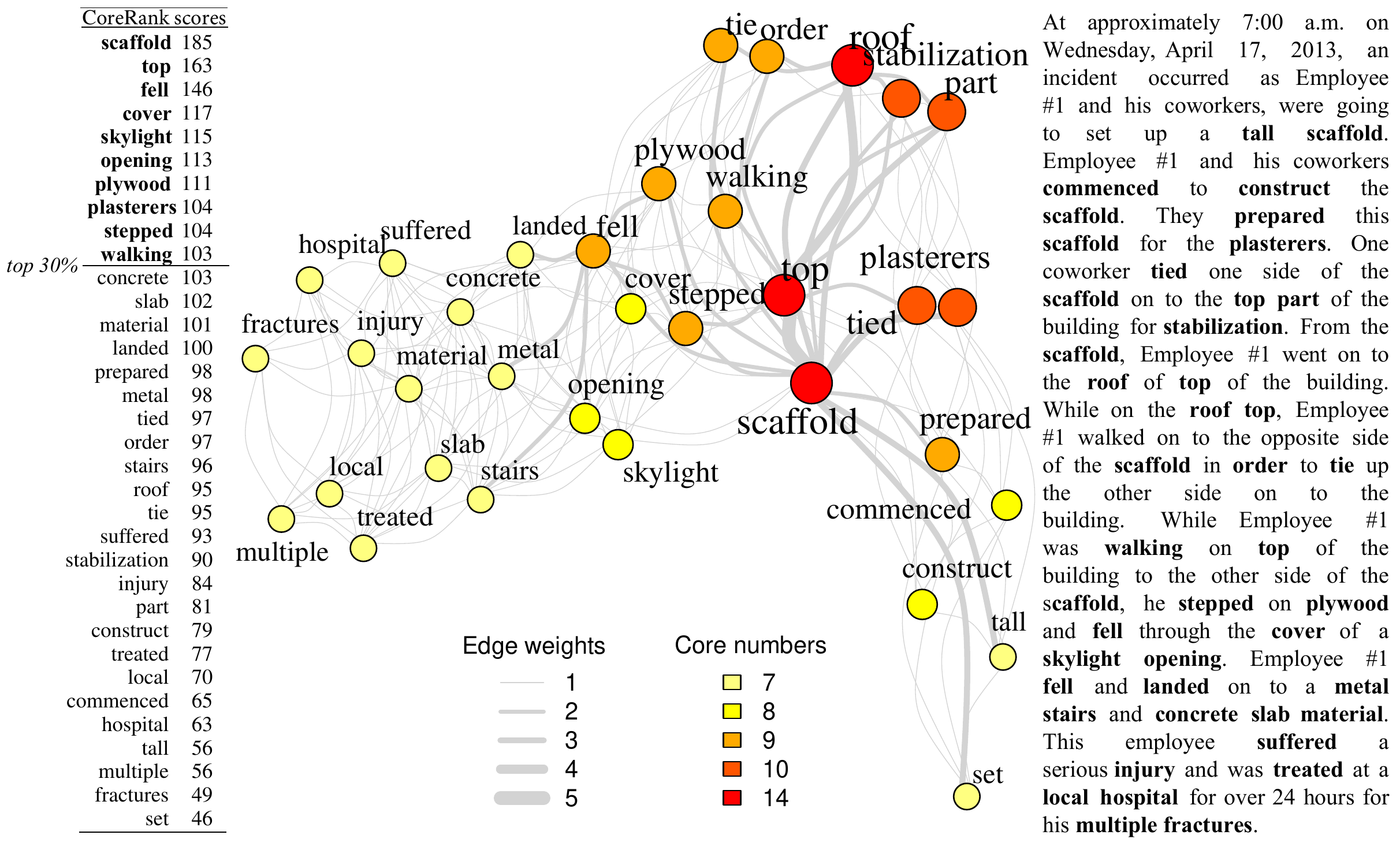}
\caption{Graph-of-words example for report ID 950613, with a sliding window of size 8. Keywords are extracted based on the CoreRank technique \cite{tixier2016degeneracy}. The CoreRank score of a node is the sum of the core numbers of its neighbors. The words in \textbf{bold} on the left are the keywords extracted from the narrative, while the words in \textbf{bold} in the report on the right correspond to the words for which embeddings are available (which make the vertices of the graph-of-words).}
\label{fig:gow}
\end{figure*}

To perform keyword extraction, we used the fully unsupervised CoreRank technique of \cite{tixier2016degeneracy}. In short, and as illustrated in Figure \ref{fig:gow}, this approach builds a graph-of-words from a document, decomposes the graph into its $k$-cores, and assigns a score to each node that corresponds to the sum of the core numbers of its neighbors. Finally, the top $p \%$ nodes are retained as keywords. Here, we used $p=30\%$.

As illustrated in Figure \ref{fig:gow}, a graph-of-words represents a piece of text as a graph where nodes are unique nouns and adjectives in the document, and where there is an edge between two nodes if the terms they represent co-occur  within a sliding window of predetermined size $W$ that is moved along the entire document from start to finish.

\begin{figure}[H]
\centering
\captionsetup{justification=centering,margin=0.1cm,font=small}
\includegraphics[height=4cm, width=4.5cm]{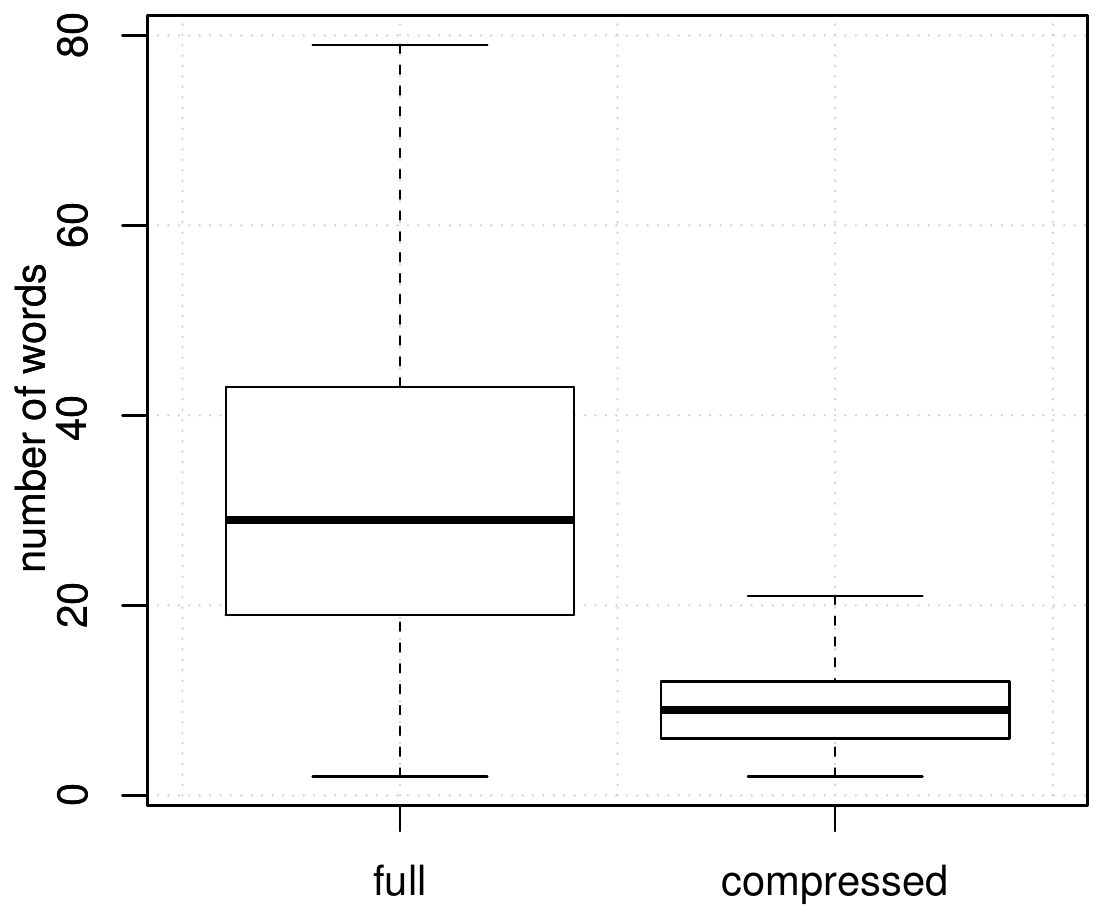}
\caption{Size (in number of words) of the full versions of the reports versus their keyword-based representation.}
\label{fig:boxplot}
\end{figure}

Graph-of-words have many graph building and mining parameters \cite{tixier2016gowvis}. Here, we weighted edges based on co-occurrence counts, and disregarded their direction (undirected graph). Finally, we ran a window of size $W=8$ over the processed versions of the reports. Note that we only performed keyword extraction for the reports featuring at least 15 words. For smaller reports, the full text was kept. The keywords for each report can be found here\footnote{\url{https://github.com/Tixierae/WECD/blob/master/compressed_reports.txt}}. The keyword extraction led to a significant compression illustrated in Figure \ref{fig:boxplot}, and to a significant speed-up too, as will be discussed in the next section.

\subsection{Results}
Classification performance using the full text of the reports is shown for each class in Tables \ref{table:res_sev}, \ref{table:res_type}, and \ref{table:res_tra}, respectively for the \textit{injury severity}, \textit{injury type}, and \textit{construction trade} prediction tasks. These results were obtained with the {\small\verb|classification_report|} function of the {\small\verb|scikit-learn|} Python library. The metric used is the standard F1-score, computed (at the class level) for each observation and then averaged (i.e., macro-averaging). A F1-score of 1 indicates perfect classification, while null values mean no skill. The support is the number of \textit{observations} (and not of \textit{predictions}) in each class. For each prediction task, we compare approaches for the number of neighbors that yielded the best absolute performance.

\textbf{Injury severity}.
For \textit{injury severity}, all methods can discriminate well \textit{hospitalized} injuries from the rest of the outcomes, and to a lesser extent, \textit{fatalities}. However, performance is very bad for \textit{non-hospitalized} injuries. 

\begin{table}[H]
\small
\centering
\begin{tabular}{r|c|c|c}
\hline
 & non-hospitalized & hospitalized & fatality \\
\hline
\hline
custom & 12.57 & 87.86 & 68.90 \\
google & 12.72 & \textbf{88.59} & \textbf{73.59}\\
bow & \textbf{16.67} & 76.65 & 56.60 \\
\hline
support & 135 & 1193 & 340\\
\hline
\end{tabular}

\captionsetup{justification=centering, size=scriptsize}
\caption{Classification performance for \textit{injury severity} (5 neighbors, macro-averaged F1 score, best score per column in \textbf{bold}).}
\label{table:res_sev}
\end{table}
For the two categories associated with good skill, the rankings are invariant: \textit{google} reaches best performance, followed by our \textit{custom} word embeddings (0.73 and 4.69 absolute improvements, respectively for \textit{hospitalized} injuries and \textit{fatalities}). This can be explained by the fact that the vocabulary that discriminates between the different severity levels is not specific to the construction industry at all. Therefore, our custom embeddings cannot compensate for the small size of their corpus of origin (11M-word) with better construction knowledge of the domain, and the \textit{google} word vectors (trained on a huge general 100B-word corpus) take the lead. 

\textbf{Injury type}.
The results are very different for \textit{injury type}. For this prediction task, our construction-specific word embeddings reach best performance for 4 categories out of 9 (with absolute improvements over \textit{google} ranging between 3.45 and 0.21), and are as good as the Google embeddings on a fifth class. 
\begin{table}[H]
\small
\centering
\begin{tabular}{r|c|c|c|c}
\hline
 & custom & google & bow & support \\
\hline
\hline
asphyxia   &  \textbf{62.79} & 59.34 & 16.84 & 43 \\
puncture & \textbf{72.13} & 71.67 & 51.49 & 70\\
amputation & 71.91 & \textbf{72.78} & 52.36 & 173\\
concussion & \textbf{24.24} & 18.97 & 27.38 & 151\\
burn & 59.46 & 59.46 & 07.14 & 26\\
cut/laceration & 37.19 & \textbf{42.36} &28.24 & 143 \\
bruise/contusion & 08.57 & \textbf{10.77} & 03.17 & 112\\
fracture & \textbf{80.56} & 80.35 & 74.22 & 900\\
electric shock & 80.46 & \textbf{81.82} & 49.32 & 50\\
\hline
\end{tabular}
\captionsetup{justification=centering, size=scriptsize}
\caption{Classification performance for \textit{injury type} (5 neighbors, macro-averaged F1-score, best score per row in \textbf{bold}).}
\label{table:res_type}
\end{table}
It is interesting to note that our embeddings tend to outperform Google ones when the absolute performance is high: of the 4 classes for which \textit{custom} reaches best performance, 3 are associated with F1-scores over 60\%, while 2 out of the 4 categories for which \textit{google} wins correspond to F1-scores below 43\%. Finally, it can be observed that the bag-of-words representation coupled with the euclidean distance leads to very poor results, which shows well the value added by the joint use of \textit{distributed representations of words} and the WMD.

\textbf{Construction trade}.
For this prediction task, our custom embeddings outperform Google ones for 2 classes out of 4. For \textit{painters}, the margin is even quite wide: using \textit{custom} leads to an absolute improvement of 7.57 in F1-score over \textit{google}. Conversely, for the categories for which \textit{google} is better, the absolute improvements do not exceed 1. 

\begin{table}[H]
\small
\centering
\begin{tabular}{r|c|c|c|c}
\hline
 & roofers & carpenters & laborers & painters \\
\hline
\hline
custom & \textbf{31.76} & 61.23 & 66.06 & \textbf{33.33} \\
google & 31.28 & \textbf{62.00} & \textbf{67.03} & 25.76 \\
bow & 29.93 & 41.48 & 61.57 & 12.73 \\
\hline
support & 134 & 573 & 867 & 94 \\
\hline
\end{tabular}
\captionsetup{justification=centering, size=scriptsize}
\caption{Classification performance for \textit{construction trade} (10 neighbors, macro-averaged F1-score, best score per column in \textbf{bold}).}
\label{table:res_tra}
\end{table}

It is also interesting to note that the 2 classes on which our word vectors reach best performance (namely \textit{roofers} and \textit{painters}) are small categories (less than 134 observations in each case, see the ``support'' row). This could mean that for those small domains, knowledge about specific construction vocabulary, brought by our local embeddings but not by the more global Google ones, is necessary to do well. For this prediction task again, \textit{bow} performs poorly (although it is surprisingly competitive for the \textit{roofers} class).

\textbf{Impact of keyword-based compression}. As can be observed from Table \ref{table:impact}, using the keywords extracted from the injury reports rather than their full text leads to a significant speed-up (8 times faster). More precisely, compute time in seconds dropped from 9.7K to 1.1K, as measured during 4-fold cross-validation on an 8-core Intel Xeon 2.4GHz machine (within-fold multiprocessing). The associated relative decrease in performance is around 10\%. It was measured in terms of overall F1-score computed over \textit{observations} (not over \textit{categories}).

\begin{table}[H]
\small
  \centering
 \begin{tabular}{r|c|c|c|c}
\hline
 & inj. severity & inj. type & trade & speed-up \\
\hline
\hline
custom & 10.68 & 10.29 & 12.00 & 8X \\
google & 10.92 & 08.75 & 12.29 & 8X \\
bow & 04.70 & 05.33 & 06.46 & NA\\
\hline
\end{tabular}
\captionsetup{justification=centering, size=scriptsize}
\caption{Relative classification performance decrease (in \%) for all prediction tasks when the compressed representation of injury reports is used rather than their full text (retaining the 30\% most highly ranked words as keywords).}
\label{table:impact}
\end{table}

\textbf{Discussion}. Overall, it should be noted that the performance of the Google News word vectors is remarkable. Even though they were not trained on a construction-specific corpus, they still manage to reach best performance in many cases. This tends to corroborate \cite{kusner2015word, mikolov2013efficient} who observed that using \textit{more} data is superior than using \textit{relevant} data when training embeddings. However, our custom word vectors do outperform Google ones in roughly half of the cases (sometimes with a wide margin), suggesting that in some applications, local embeddings are indeed better than global ones. These results are in accordance with \cite{diaz2016query}.

\section{Conclusion and Next steps}{\label{sec:conc}}
We presented one of the earliest applications of word embeddings to the construction domain. In addition to releasing one of the largest publicly available collections of raw construction-related text to date (11M words, 450K unique words), and a novel data set for injury report classification, we showed through multiple examples and a case study that the use of word embeddings in the construction industry is very promising and has many potential applications. By allowing more flexible, semi-supervised classification of injury reports into categories, they could be used to better predict and prevent injuries \cite{tixier2016application} or for more accurate safety risk modeling and simulation \cite{tixier2016construction}.

We could improve our embeddings in several ways. 
First, our 11M-word corpus can obviously be augmented. Even though we tried to be comprehensive, there are surely many large freely accessible sources of construction text that we did not leverage. This is one of the reasons why we decided to release our corpus as publicly available, so that others can add to it and let the entire construction research community benefit from these additions. Second, we did not tune any of the parameters of {\small\verb|word2vec|} (architecture, window size, downsampling threshold, negative sampling value...), mainly because we were more interested in providing an initial motivation for the use of word embeddings in the construction domain rather than reaching best possible performance. We could verify that out-of-the-box, embeddings learned with standard parameter values do well qualitatively and quantitatively, but fine-tuning could certainly yield better performance. 

However, we did quickly check that embedding phrases (in our case, only bigrams) instead of simple unigrams was not giving better results. But once again, this may depend on the parameter values used. We also evaluated embeddings trained solely on the injury reports sub-corpora rather than on the full corpus, but performance was not better. Using pre-trained word vectors (such as Google News) as a starting point and continuing training on our corpus could be worth investigating.

Finally, it seems that keyword-based compression can significantly increase distance computation speed with only a limited drop in accuracy. We used the CoreRank technique out-of-the-box for exploration purposes, but did not perform any kind of parameter tuning. In the end, the WMD is mainly determined by a few highly discriminative words in each document. Therefore, if these words could happen to be the extracted keywords (through tuning), the decrease in performance would be only marginal. Future work should also investigate the trade-off between compression ratio (number of keywords retained) and classification performance.

\bibliography{reference}

\begin{thebibliography}{16}
\providecommand{\natexlab}[1]{#1}
\providecommand{\url}[1]{\texttt{#1}}
\expandafter\ifx\csname urlstyle\endcsname\relax
  \providecommand{\doi}[1]{doi: #1}\else
  \providecommand{\doi}{doi: \begingroup \urlstyle{rm}\Url}\fi

\bibitem[Bengio et al.(2003)]{bengio2003neural}
Bengio, Yoshua, Ducharme, R{\'e}jean, Vincent, Pascal, and Jauvin, Christian.
\newblock A neural probabilistic language model.
\newblock \emph{Journal of Machine Learning Research},
 \penalty0 (3):\penalty0 1137--1155, 2003.

\bibitem[Caldas and Soibelman(2003)]{caldas2003automating}
Caldas, Carlos H., and Soibelman, Lucio.
\newblock Automating hierarchical document classification for construction management information systems.
\newblock \emph{Automation in Construction},
 4\penalty0 :\penalty0 395--406, 2003.

\bibitem[Chokor et al.(2016)]{chokor2016analyzing}
Chokor, Abbas, Naganathan, Hariharan, Chong, Wai K., and El Asmar, Mounir
\newblock Analyzing Arizona OSHA Injury Reports Using Unsupervised Machine Learning.
\newblock \emph{Procedia Engineering},
 145\penalty0 :\penalty0 1588--1593, 2016.

\bibitem[Cover and Hart(1967)]{cover1967nearest}
Cover, Thomas and Hart, Peter.
\newblock Nearest neighbor pattern classification.
\newblock \emph{IEEE transactions on information theory},
 13\penalty0 :\penalty0 (1) 21--27, 1967.

\bibitem[Diaz et al.(2016)]{diaz2016query}
Diaz, Fernando, Mitra, Bhaskar, and Craswell, Nick.
\newblock Query Expansion with Locally-Trained Word Embeddings.
\newblock \emph{arXiv preprint arXiv:1605.07891}
,2016

\bibitem[Harris(1954)]{harris1954distributional}
Harris, Zellig S.
\newblock Distributional structure.
\newblock \emph{Word},
 10\penalty0 (2-3):\penalty0 146--162, 1954.

\bibitem[Katz(1987)]{katz1987estimation}
Katz, Slava.
\newblock Estimation of probabilities from sparse data for the language model component of a speech recognizer.
\newblock \emph{IEEE transactions on acoustics, speech, and signal processing},
 35\penalty0 (3):\penalty0 400--401, 1987.
 
\bibitem[Kusner et al.(2015)]{kusner2015word}
Kusner, Matt J, Sun, Yu, Kolkin, Nicholas, and Weinberger, Kilian Q.
\newblock From word embeddings to document distances.
\newblock In \emph{Proceedings of the 32nd International Conference on Machine Learning (ICML 2015)}
,957--966, 2015.
  
\bibitem[Mikolov et al.(2013a)]{mikolov2013efficient}
Mikolov, Tomas, Chen, Kai, Corrado, Greg, and Dean, Jeffrey.
\newblock Efficient estimation of word representations in vector space.
\newblock \emph{arXiv preprint arXiv:1301.3781}, 2013a.

\bibitem[Mikolov et al.(2013b)]{mikolov2013exploiting}
Mikolov, Tomas, Le, Quoc V., and Sutskever, Ilya.
\newblock Exploiting similarities among languages for machine translation.
\newblock \emph{arXiv preprint arXiv:1309.4168},
2013b.

\bibitem[Mikolov et al.(2013c)]{mikolov2013linguistic}
Mikolov, Tomas, Yih, Wen-tau, and Zweig, Geoffrey.
\newblock Linguistic Regularities in Continuous Space Word Representations.
\newblock In \emph{Proceedings of HLT-NAACL}
,746--751, 2013c.

\bibitem[Mikolov et al.(2013d)]{mikolov2013distributed}
Mikolov, Tomas, Sutskever, Ilya, Chen, Kai, Corrado, Greg S, and Dean, Jeff.
\newblock Distributed representations of words and phrases and their compositionality.
\newblock \emph{Advances in neural information processing systems},
,3111--3119, 2013d.
 
\bibitem[Pedregosa et al.(2011)]{scikit-learn}
Pedregosa, F., Varoquaux, G., Gramfort, A., Michel, V., Thirion, B., Grisel, O., Blondel, M. Prettenhofer, P., Weiss, R., Dubourg, V., Vanderplas, J., Passos, A., Cournapeau, D., Brucher, M., Perrot, M., and Duchesnay, E.
\newblock Scikit-learn: Machine Learning in {P}ython.
\newblock \emph{Journal of Machine Learning Research},
 12\penalty0 :\penalty0 2825--2830, 2011.

\bibitem[R Core Team(2015)]{r}
R Core Team.
\newblock R: A Language and Environment for Statistical Computing.
\newblock \emph{R Foundation for Statistical Computing}, 2015.

\bibitem[{\v R}eh{\r u}{\v r}ek and Sojka(2010)]{gensim}
Radim {\v R}eh{\r u}{\v r}ek and Petr Sojka.
\newblock Software Framework for Topic Modelling with Large Corpora.
\newblock In \emph{Proceedings of the LREC 2010 Workshop on New Challenges for NLP Frameworks}
, 45--50, 2010.

\bibitem[Rong(2014)]{rong2014word2vec}
Rong, Xin.
\newblock word2vec parameter learning explained.
\newblock \emph{arXiv preprint arXiv:1411.2738},
2014.

\bibitem[Rubner et al.(2000)]{rubner2000earth}
Rubner, Yossi and Tomasi, Carlo and Guibas, Leonidas J.
\newblock The earth mover's distance as a metric for image retrieval.
\newblock \emph{International journal of computer vision}
, 99--121, 2000.

\bibitem[Salton and Buckley(1988)]{salton1988term}
Salton, Gerard, and Buckley, Christopher.
\newblock Term-weighting approaches in automatic text retrieval.
\newblock \emph{Information processing \& management},
 5\penalty0 :\penalty0 513--523, 1988.

\bibitem[Tixier et al.(2016a)]{tixier2016automated}
Tixier, Antoine J.-P., Hallowell, Matthew R., Rajagopalan, Balaji, and Bowman, Dean.
\newblock Automated content analysis for construction safety: A natural language processing system to extract precursors and outcomes from unstructured injury reports.
\newblock \emph{Automation in Construction},
62\penalty0 :\penalty0 45--56, 2016a.

\bibitem[Tixier et al.(2016b)]{tixier2016degeneracy}
Tixier, Antoine J.-P., Malliaros, Fragkiskos. D., and Vazirgiannis, Michalis. 
\newblock A Graph Degeneracy-based Approach to Keyword Extraction.
\newblock In \emph{Proceedings of EMNLP},
2016b.

\bibitem[Tixier et al.(2016c)]{tixier2016gowvis}
Tixier, Antoine J.-P., Skianis, Konstantinos, and Vazirgiannis, Michalis.
\newblock GoWvis: a web application for Graph-of-Words-based text visualization and summarization.
\newblock In \emph{Proceedings of ACL}
, paper 151, 2016c.

\bibitem[Tixier et al.(2016d)]{tixier2016application}
Tixier, Antoine J.-P., Hallowell, Matthew R., Rajagopalan, Balaji, and Bowman, Dean.
\newblock Application of machine learning to construction injury prediction.
\newblock \emph{Automation in Construction},
 69\penalty0 :\penalty0 102--114, 2016d.

\bibitem[Tixier et al.(2016e)]{tixier2016construction}
Tixier, Antoine J.-P., Hallowell, Matthew R., and Rajagopalan, Balaji.
\newblock Construction Safety Risk Modeling and Simulation.
\newblock \emph{arXiv preprint arXiv:1609.07912},
2016e.

\bibitem[Williams and Gong(2014)]{williams2014predicting}
Williams, Trefor P., and Gong, Jie.
\newblock Predicting construction cost overruns using text mining, numerical data and ensemble classifiers.
\newblock \emph{Automation in Construction},
 43\penalty0 :\penalty0 23--29, 2014.

\bibitem[Yu and Hsu(2013)]{hsu2013content}
Yu, Wen-der, Hsu, and Jia-yang.
\newblock Content-based text mining technique for retrieval of CAD documents.
\newblock \emph{Automation in Construction},
 31\penalty0 :\penalty0 65--74, 2013.

\end{thebibliography}
\bibliographystyle{icml2016}
\end{document}